\documentclass[letterpaper, 10 pt, conference]{ieeeconf}  

\IEEEoverridecommandlockouts                              

\usepackage{graphics} 
\usepackage{epsfig} 
\usepackage{amsmath} 
\usepackage{amssymb}  
\usepackage[hidelinks]{hyperref}
\usepackage{lipsum}
\usepackage{tikz}
\usepackage{float}
\usetikzlibrary{tikzmark}
\usepackage{array}
\newcolumntype{L}{>{\centering\arraybackslash}m{0.7cm}}
\graphicspath{{images/}}

\usepackage[caption=false, font=footnotesize]{subfig}
\title{\LARGE \bf
Single Network Panoptic Segmentation for Street Scene Understanding}

\author{Daan de Geus, Panagiotis Meletis, and Gijs Dubbelman
\thanks{Daan de Geus ({\tt\small \href{mailto:d.c.d.geus@tue.nl}{d.c.d.geus@tue.nl}}), Panagiotis Meletis ({\tt\small \href{mailto:p.c.meletis@tue.nl}{p.c.meletis@tue.nl}}) and Gijs Dubbelman
({\tt\small \href{mailto:g.dubbelman@tue.nl}{g.dubbelman@tue.nl}}) are with the Department of Electrical Engineering,
Eindhoven University of Technology, Eindhoven, The Netherlands.}%
}

\begin{document}

\maketitle
\thispagestyle{empty}
\pagestyle{empty}

\begin{abstract}
In this work, we propose a single deep neural network for panoptic segmentation, for which the goal is to provide each individual pixel of an input image with a class label, as in semantic segmentation, as well as a unique identifier for specific objects in an image, following instance segmentation. Our network makes joint semantic and instance segmentation predictions and combines these to form an output in the panoptic format. This has two main benefits: firstly, the entire panoptic prediction is made in one pass, reducing the required computation time and resources; secondly, by learning the tasks jointly, information is shared between the two tasks, thereby improving performance. Our network is evaluated on two street scene datasets: Cityscapes and Mapillary Vistas. By leveraging information exchange and improving the merging heuristics, we increase the performance of the single network, and achieve a score of 23.9 on the Panoptic Quality (PQ) metric on Mapillary Vistas validation, with an input resolution of 640 x 900 pixels. On Cityscapes validation, our method achieves a PQ score of 45.9 with an input resolution of 512 x 1024 pixels. Moreover, our method decreases the prediction time by a factor of 2 with respect to separate networks.

\end{abstract}

\section{INTRODUCTION}

Scene understanding plays a crucial role in automated driving, and image recognition provides a way to achieve this. The main goal for image recognition is to identify all elements in an image. At a high level, these elements can be divided into two categories: \textit{stuff} and \textit{things} classes \cite{Forsyth1996}. \textit{Things} are countable objects, such as vehicles, persons and traffic signs. On the other hand, \textit{stuff} is the set of remaining elements, usually not countable, such as sky, road and water. 

\begin{figure}[t]
\centering
\includegraphics[width=0.49\linewidth]{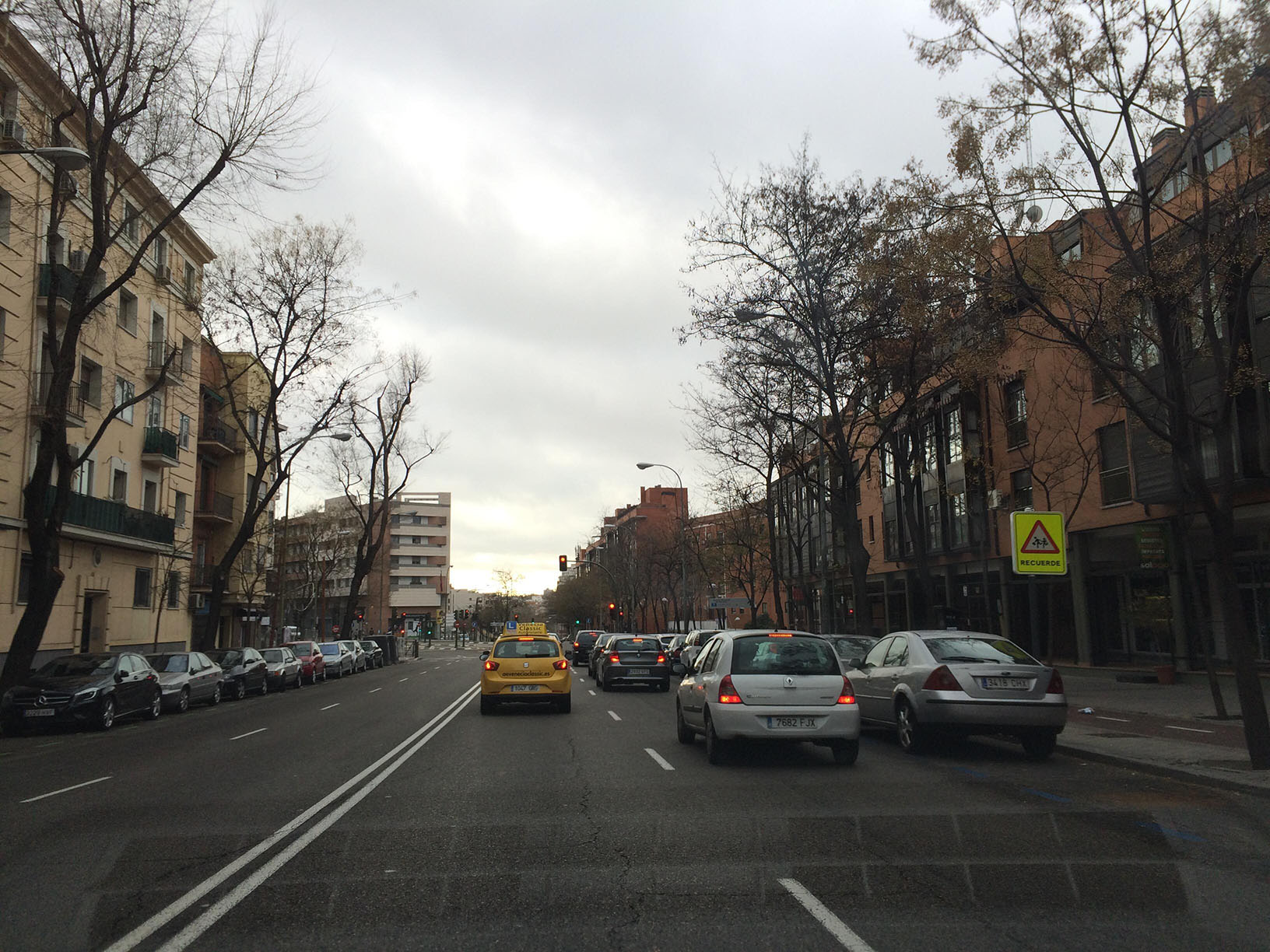}
\includegraphics[width=0.49\linewidth]{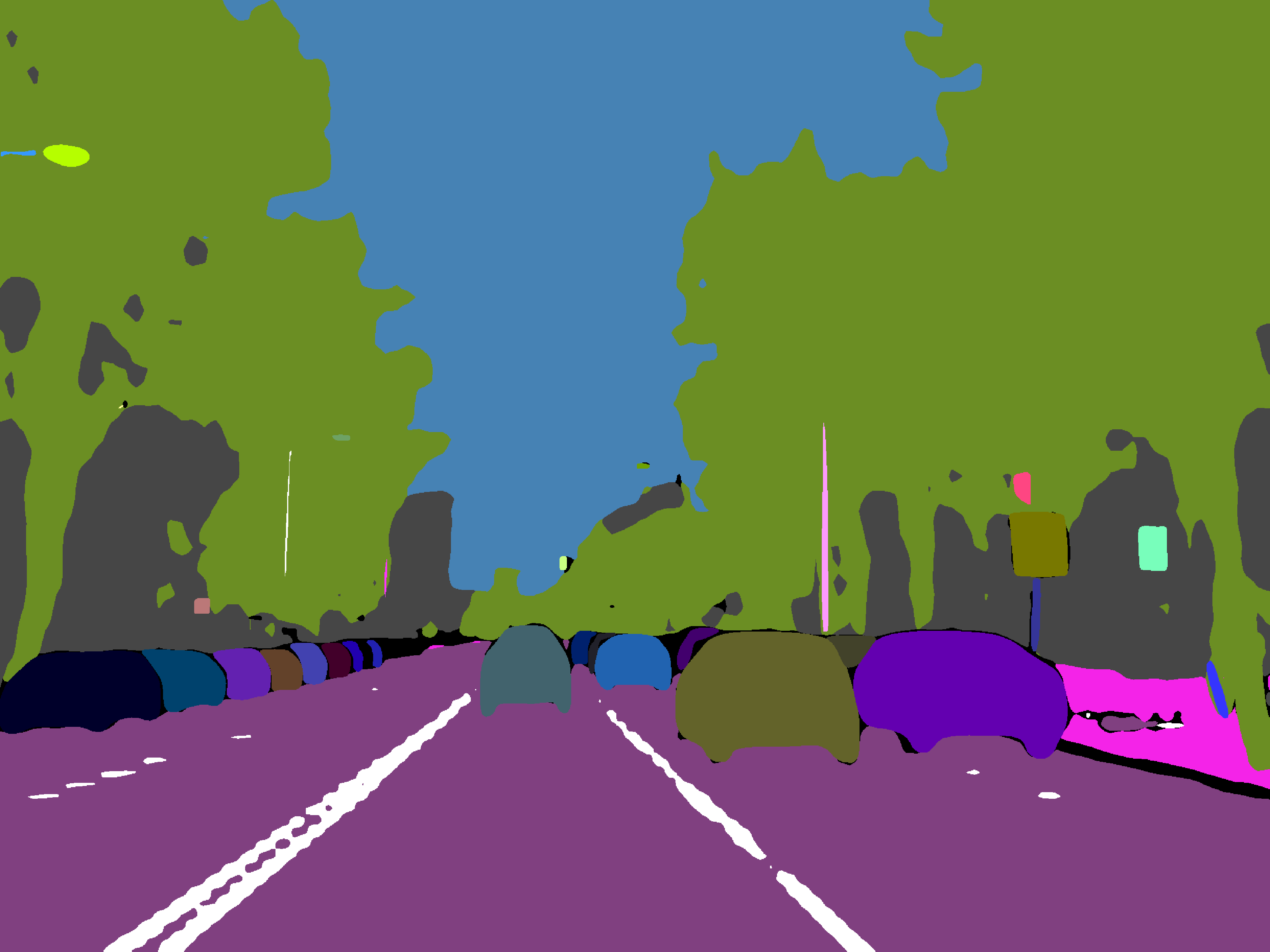}\\
\includegraphics[width=0.49\linewidth]{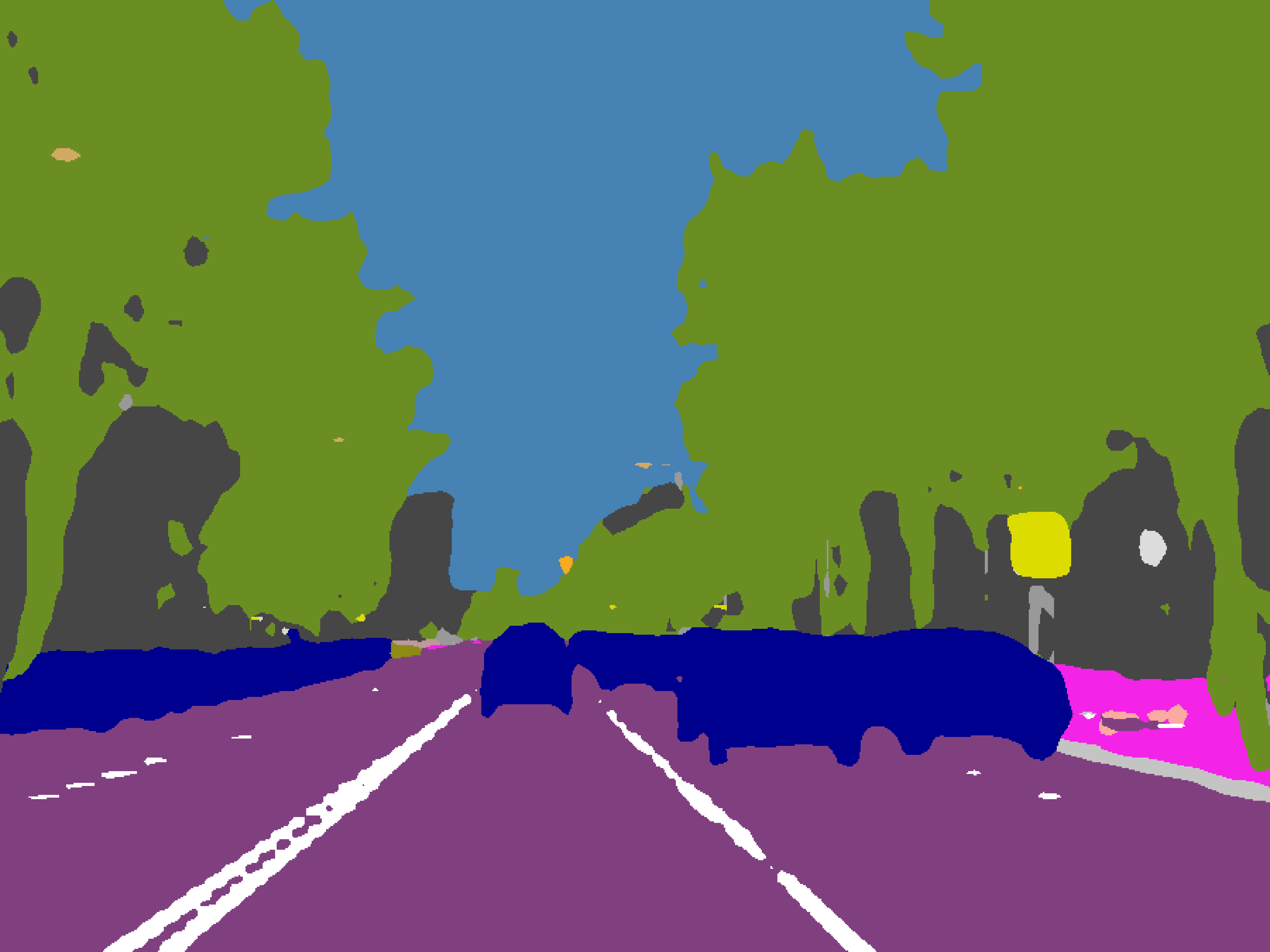}
\includegraphics[width=0.49\linewidth]{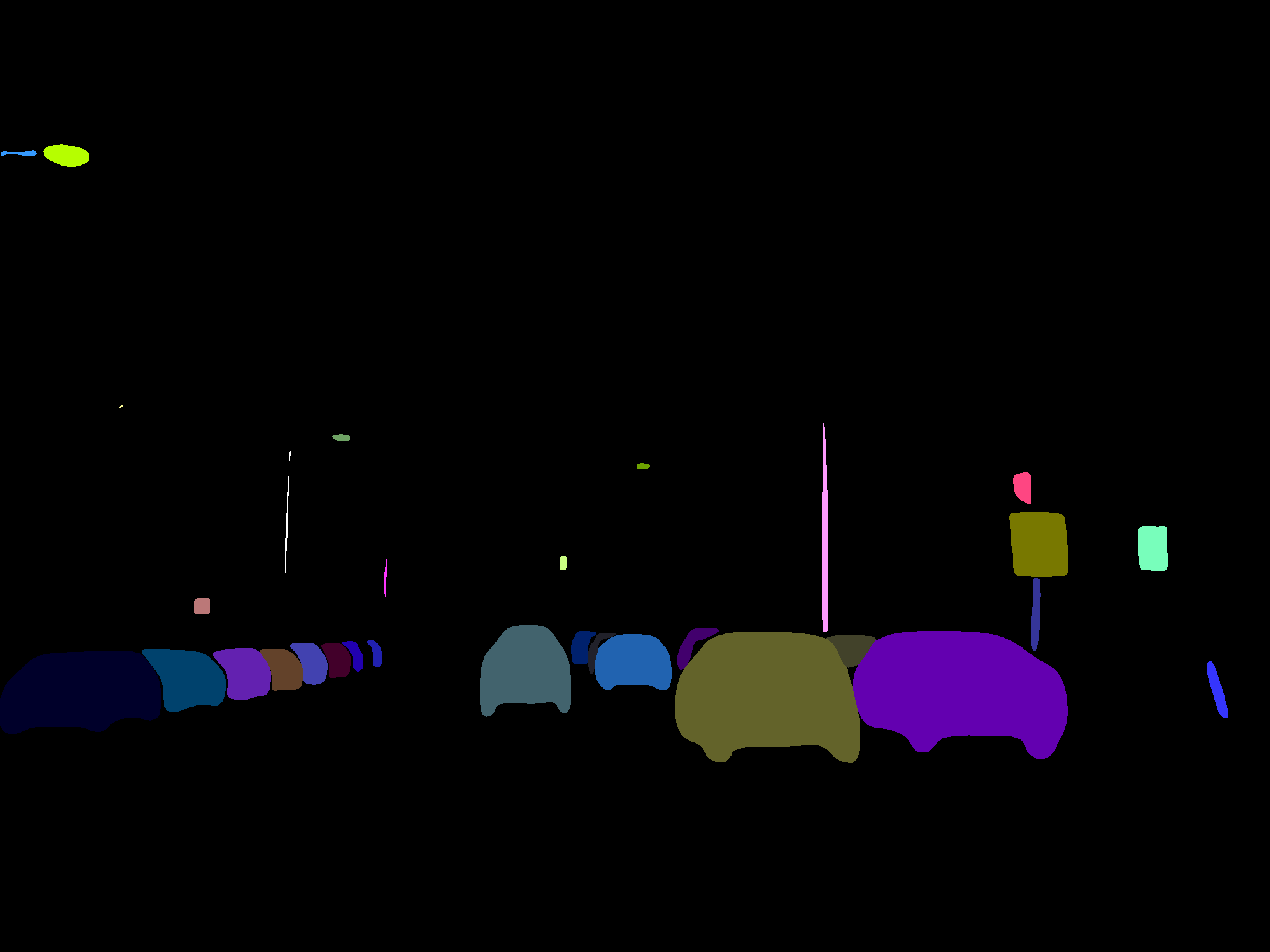}
\caption{A panoptic segmentation prediction by the network. Top left: original input image, from the Mapillary Vistas validation set. Top right: panoptic segmentation prediction by our system. Each pixel receives a class label and all pixels belonging to specific objects also receive a unique identifier. Bottom left: semantic segmentation prediction, where all pixels only receive a class label. Bottom right: instance segmentation prediction, where only pixels of specific object classes receive a class label and an identifier label.}
\label{fig:eye_catcher}
\end{figure}

Instance segmentation and semantic segmentation are two very important image recognition tasks. Both aim at describing the content of an image as detailed as possible, and approach this in two different ways. The first task, instance segmentation, focuses on the detection and segmentation of \textit{things}. If an object is detected, a pixel mask is predicted for this object, and the output of such a method is a set of pixel masks (see Fig. \ref{fig:eye_catcher}, bottom right). By design, this method does not account for all elements in an image, as it does not consider \textit{stuff} classes. The second task, semantic segmentation, does consider all elements, as the aim is to make a class prediction for each pixel in an image, for both \textit{things} and \textit{stuff} classes. However, the semantic segmentation output does not differentiate between different instances of \textit{things} (see Fig. \ref{fig:eye_catcher}, bottom left). As a result, both methods lack the ability to fully describe the contents of an image.

To bridge this gap, the task of \textit{panoptic segmentation} has recently been introduced \cite{Kirillov2018}. For panoptic segmentation, the goal is to predict 1) a class label and 2) an instance \textit{id} for all pixels in an image. This instance \textit{id} is used to differentiate between different object instances; all pixels with the same instance \textit{id} belong to the same object. By definition, all \textit{stuff} predictions of the same class receive the same instance \textit{id}. An example is given in Fig. \ref{fig:eye_catcher} (top right). In \cite{Kirillov2018}, a baseline method is proposed that fuses the output of separate state-of-the-art semantic segmentation and instance segmentation networks using basic heuristics. This allows for use of models that are optimal for both individual tasks, but this means that there is no single network. A single network is desirable because it allows for easier implementation on devices, and it can significantly decrease the computational time and resources required to make a prediction, which is very relevant for application in intelligent vehicles.

\begin{figure*}[t]
\centering
\includegraphics[width=\linewidth]{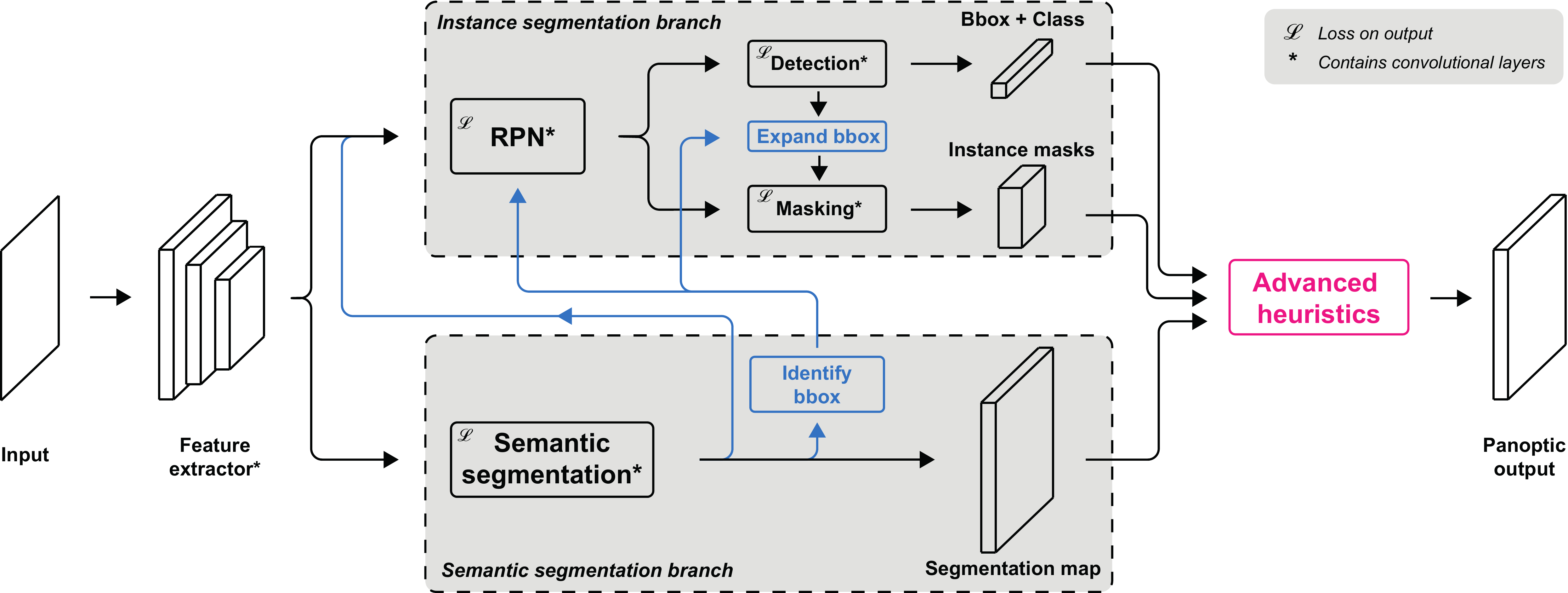}
\caption{Our single network architecture for panoptic segmentation. The network consists of an instance segmentation branch and semantic segmentation branch that share the same feature extractor. We introduce information exchange between the branches to improve the performance. The additional information flow is indicated in blue, and explained in Section \ref{sec:method_info_exchange}. Finally, the outputs of both branches are merged using advanced heuristics to form a panoptic output, as indicated in purple (see Section \ref{sec:method_heuristics} and Fig. \ref{fig:heuristic_overview}).}
\label{fig:baseline_architecture}
\end{figure*}

In this work, therefore, we research and present a single deep neural network for panoptic segmentation. This network consists of a common feature extractor and two different branches that output semantic segmentation and instance segmentation predictions. This joint network architecture leads both to conflicts and opportunities, which are both addressed by leveraging the most optimal information from both branches of the network. To get a final consistent panoptic segmentation output, the semantic segmentation and instance segmentation outputs are fused using advanced heuristics.

To summarize, our main contributions to street scene understanding from image data are:
\begin{itemize}
\item A single network for panoptic segmentation.
\item Inter-branch information exchange to leverage the single network architecture.
\item Improved heuristics for merging the semantic and instance segmentation predictions.
\end{itemize}

The implementation of our network is made available to the research community \cite{CodeGitlab2019}. Preliminary results were submitted to the COCO \& Mapillary Joint Recognition Challenge at ECCV 2018 \cite{DeGeus2018}.

In the remainder of this paper, we will first review the related literature in Section \ref{sec:related_work}. Thereafter, in Section \ref{sec:methodology}, we discuss our methodology. Subsequently, in  Section \ref{sec:implementation_details} the implementation details of our experiments are provided. The results on these experiments are presented in Section \ref{sec:results}. Finally, we provide conclusions in Section \ref{sec:conclusions}.

\section{RELATED WORK}
\label{sec:related_work}
The task of panoptic segmentation is closely related to semantic segmentation and instance segmentation. Both these tasks have seen great progress over the last years.

In semantic segmentation, it is very important that spatial relations are preserved, since the output is directly spatially related to the input. For this reason, the application of convolutional layers is essential.  The first semantic segmentation architecture that consists of a Fully Convolutional Network (FCN), i.e. applying only convolutional layers, was presented in \cite{Shelhamer2017}. They apply an FCN to decode the image into feature maps, make class predictions on these feature maps, and apply bilinear upsampling to create the segmentation masks. The SegNet model \cite{Badrinarayanan2017} is also an FCN, but it applies a decoding network instead of bilinear upsampling. As of recently, PSPNet is the state-of-the-art model, as it improves performance by leveraging information from different levels of the feature map, introducing a sense of context \cite{Zhao2017}.

Instance segmentation, on the other hand, is closely related to bounding box object detection. Instance segmentation extends object detection by predicting per-pixel masks for the detected objects. Therefore, many methods choose to make instance segmentation predictions by predicting instance masks for detected objects. A state-of-the art instance segmentation method is Mask R-CNN \cite{He2017}. In this approach, the object detection method of Faster R-CNN \cite{Ren2017} is extended with per-pixel instance mask predictions for for each bounding box that is likely to contain an object. Recently, the Mask R-CNN architecture has been improved with the development of Feature Pyramid Networks \cite{Lin2017} and the Path Aggregation Network \cite{Liu2018}, leading to new state-of-the-art results.

We have seen that, so far, separate instance segmentation and semantic segmentation networks have been used for panoptic segmentation \cite{Kirillov2018}. As a result, it was possible to use networks that are optimized for these specific tasks. However, there are also downsides to this method. If the predictions were made using a single network, computation time and resources could be decreased, because fewer parameters would be required. This is the case since a significant part of the processing is spent on low-level feature extraction layers that can be shared between different branches in a network. Moreover, jointly learning multiple tasks has the potential of improving performance, because information can be shared between different parts of the network. Therefore, we propose to address the task of panoptic segmentation by using a single network that makes parallel semantic segmentation and instance segmentation predictions, and fuses these outputs using heuristics. 

Furthermore, we leverage the single network architecture by introducing additional information flow within the network, to enhance the overall performance of the model. In \cite{Mao2017} and \cite{Shrivastava2016}, it has been shown that additional information flow between different tasks can improve the performance of the individual subtasks. In our network, it should improve the performance of the network as a whole.

Concurrent work also focusses on a unified single network for panoptic segmentation. In \cite{Li2018a}, the method consists of a unified network similar to ours, as well as a consistency loss to make the output more consistent, but there is no additional information flow to boost the performance. AUNet \cite{Li2018b} does leverage information exchange, but it requires complicated attention and masking operations. Our framework is designed to be simple and generally applicable, while leveraging the architecture by using additional information flow to improve the performance. The increase in related concurrent work highlights the relevance of creating a single unified network for panoptic segmentation.

\section{METHODOLOGY}
\label{sec:methodology}
We propose a panoptic segmentation method that consists of three parts: a single network architecture, inter-branch information exchange to leverage this single network architecture, and advanced heuristics to fuse the outputs. The resulting network architecture is depicted in Fig. \ref{fig:baseline_architecture}.

\subsection{Single network architecture}
Our architecture jointly makes semantic segmentation and instance segmentation predictions in a single network. This network consists of a semantic segmentation and instance segmentation branch both using the same feature extractor. These branches are trained jointly and output their predictions in one pass.

In our baseline network, we use a ResNet-50 \cite{He2015} feature extractor with an output stride of 8. The original stride of ResNet-50 is 32, but in our network it is reduced to allow for denser semantic segmentation predictions \cite{Shelhamer2017}. 

For the semantic segmentation branch, we follow \cite{Shelhamer2017}. In the original implementation, the predictions are made directly after the feature extractor. In our network, this feature extractor is shared with the instance segmentation branch. This means that there are only very few parameters that are used only for the semantic segmentation task. This could lead to decreased performance. For this reason, we add a Pyramid Pooling Module (PPM) \cite{Zhao2017} to the semantic segmentation branch. This PPM is introduced as a general improvement to semantic segmentation, but in our network it also acts as an adaptation network. Finally, to generate the final output of this branch, we apply hybrid upsampling to reshape the predictions to the size of the input image \cite{Meletis2018}. This hybrid upsampling technique first applies a learnable deconvolution operation and then bilinearly resizes the predictions to the dimensions of the input image. 

The instance segmentation branch is based on Mask R-CNN \cite{He2017}. First, a Region Proposal Network (RPN) is used to generate region proposals for potential objects in the image. The features corresponding to these proposals are then extracted from the feature map and subjected to the convolutional layers of the final ResNet-50 block. Finally, these features are used to make three different predictions for each region proposal: a classification score, bounding box coordinates, and an instance mask. After applying non-maximum suppression, the output of this branch is a set of detected objects consisting of  class, bounding box and per-pixel mask predictions.

To enable joint learning for this network, a single loss function is formed. This means that the various loss functions from the different network branches have to be combined and balanced. The total loss, $L_{tot}$, is given by

\begin{equation}
\label{eq:loss}
\begin{aligned}
L_{tot} &= {\lambda_1}{L_{rpn,obj}} + {\lambda_2}{L_{rpn,reg}} \\ &+
{\lambda_3}{L_{det,cls}} + {\lambda_4}{L_{det,reg}} \\ &+
{\lambda_5}{L_{mask}} \\ &+ {\lambda_6}{L_{seg}} \\ &+  {\lambda_7}{R}.
\end{aligned}
\end{equation}

Here, $L_{rpn,obj}$ is the softmax cross-entropy objectness loss function for the RPN, $L_{rpn,reg}$ is the smooth L1 regression loss function for the RPN \cite{Girshick2015}, $L_{det,cls}$ is the softmax cross-entropy classification loss function for object detection, $L_{det,reg}$ is the smooth L1 regression loss function for the object bounding boxes, $L_{mask}$ is the sigmoid cross-entropy loss on the instance masks, and $L_{seg}$ is the sparse softmax cross-entropy segmentation loss on the semantic segmentation outputs. Finally, $R$ is the L2 regularization on the model parameters. The weights $\lambda_1...\lambda_n$ are the $n$ tuning parameters that are used to balance the losses. The values used for these parameters are discussed in Section \ref{sec:implementation_details} and provided in Table \ref{tab:loss_weights}.

\subsection{Inter-branch information exchange}
\label{sec:method_info_exchange}

\begin{figure}[t]
\centering
\includegraphics[width=\linewidth]{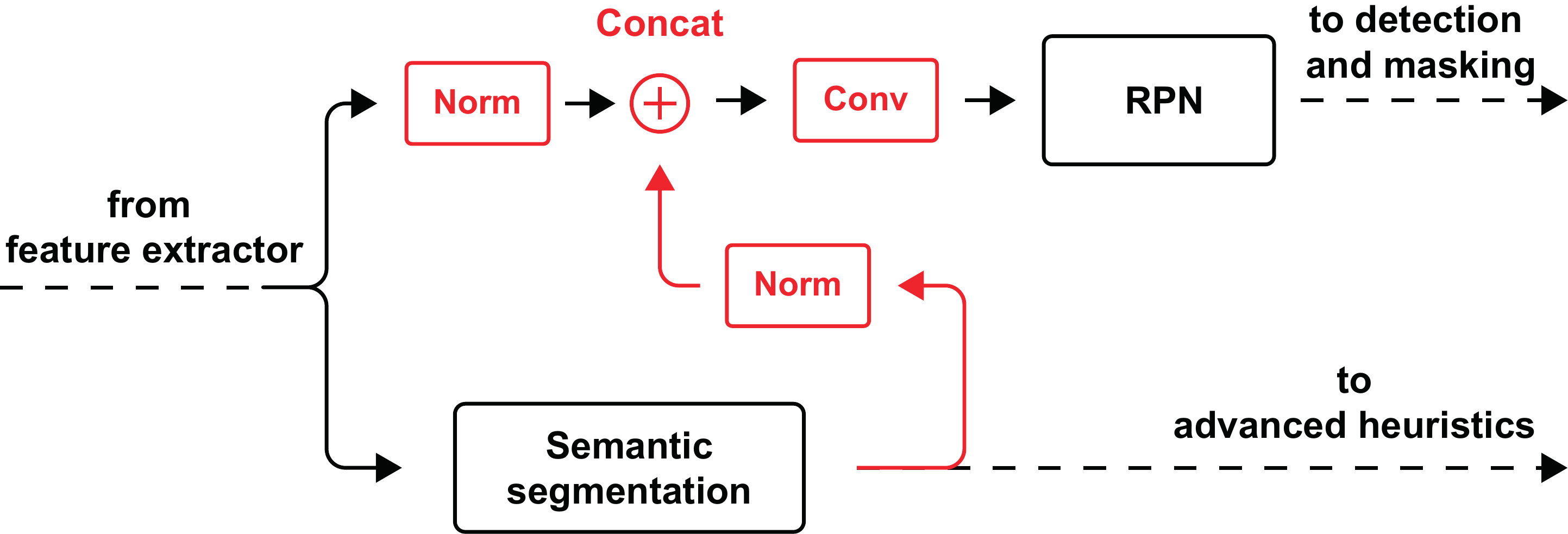}
\caption{The additional information flow for implicit information exchange. The added flow and components are indicated in red. \textit{Norm} and \textit{Concat} represent normalization and concatenation operations, respectively. \textit{Conv} is a 3x3 convolutional layer.}
\label{fig:info_exchange}
\end{figure}

Our single network architecture for panoptic segmentation introduces several opportunities over the use of separate networks. Firstly, jointly learning the semantic and instance segmentation tasks can improve the performance of both tasks, because the tasks might require similar features, which they can both retrieve and influence using the shared feature extractor. Secondly, the architecture allows to introduce additional information flow between the two semantic segmentation and instance segmentation branch; we do this in multiple basic but effective ways.

\subsubsection{Explicit information}
Certain \textit{things} predictions from the semantic segmentation branch are better than the predictions by the instance segmentation branch. Since the final output only contains \textit{things} predictions from the instance segmentation branch, potentially valuable information is lost, leading to a lower performance. To compensate for this, we use the \textit{things} predictions by the semantic segmentations to improve the instance segmentation output, in two different ways.

Firstly, we add bounding boxes to the region proposals generated by the RPN, based on the semantic segmentation output. We identify all \textit{things} clusters in the semantic segmentation output, generate bounding boxes for these clusters, and use them as additional region proposals. Secondly, we expand bounding boxes predicted by the detection branch based on the semantic segmentation output. We match all predicted bounding boxes with the corresponding \textit{things} class in the semantic segmentation output, and expand the box if the matched segment extends beyond the boundary of the box.
 
\subsubsection{Implicit information}
As became clear from \cite{Mao2017} and \cite{Shrivastava2016}, it can be beneficial to implicitly use semantic segmentation information to improve instance segmentation as well. In our network, we follow part of the method proposed by \cite{Shrivastava2016} and introduce a very basic additional information channel. We use the output from the semantic segmentation branch before the final softmax layer, normalize it and concatenate it to the normalized features from the feature map. We then apply a 3x3 convolutional layer and use the output from this layer as input to the instance segmentation branch. By doing so, we can improve the performance of both the semantic and instance segmentation branch, because the forward and backward pass through the network allow for relevant data from one branch to flow through the other. The additional information flow is depicted in Fig. \ref{fig:info_exchange}.

\subsection{Advanced merging heuristics}
\label{sec:method_heuristics}

\begin{figure}[t]
\centering
\includegraphics[width=\linewidth]{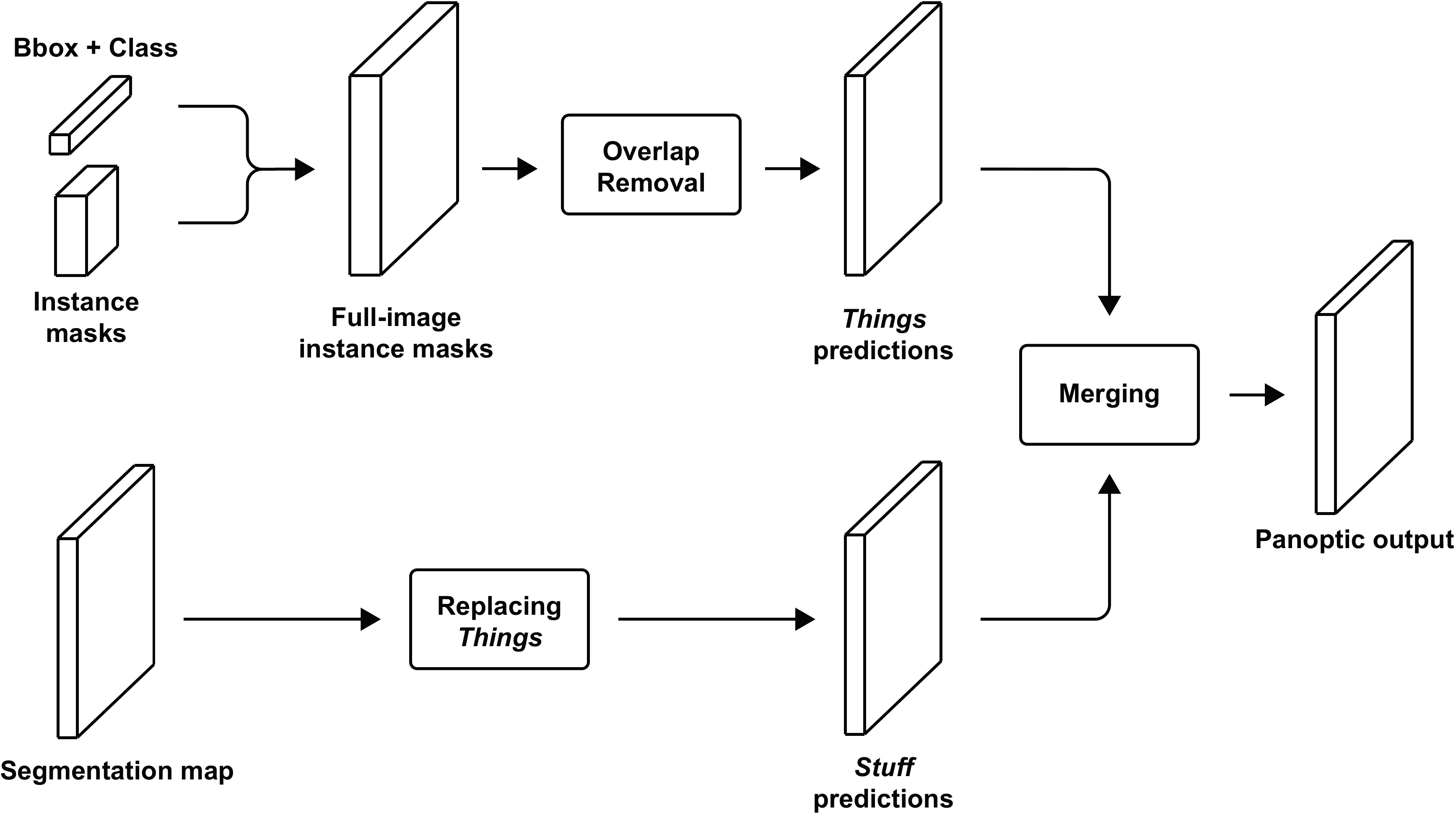}
\caption{An overview of the heuristics used for merging the instance segmentation and semantic segmentation predictions. On the top branch, we first transform the instance segmentation predictions to generate full-image instance masks. Then, we remove overlap to get a single \textit{things} prediction for each pixel. On the bottom branch, we replace the \textit{things} predictions and end up with \textit{stuff} predictions only. Finally, we generate the panoptic output by overlaying the \textit{stuff} predictions with the \textit{things} predictions.}
\label{fig:heuristic_overview}
\end{figure}

Because our network outputs two separate predictions in parallel, these outputs have to be processed in order to generate a panoptic segmentation prediction. For panoptic segmentation, two values have to be predicted for each pixel: a class label and an instance \textit{id}. There are essentially two conflicts that need to be solved before being able to generate this output: overlapping instance masks, and conflicting predictions for \textit{things} classes by the two branches. In addition to this, we apply a heuristic that removes unlikely \textit{stuff} predictions. An overview of the merging heuristics is shown in Fig. \ref{fig:heuristic_overview}.

\begin{table*}[t]
\caption{ The overall results of our method on the Mapillary Vistas validation set.}
\label{tab:results_mapillary_overall}
\centering
\begin{tabular}{ l r@{\hspace{2px}}|| c | c | c || c | c ||| c }
Method &  & PQ & SQ & RQ & PQ\textsubscript{Th} & PQ\textsubscript{St} & Prediction time\\
\hline
Multiple networks & \hspace{20px} & 21.3 & 65.2 & 25.3 & 13.8 & 31.4 & 903 ms\\
Single network & \tikzmark{b}& 21.0 & 65.6 & 27.5 & 14.9 & 29.2 & 532 ms \\
+ Advanced heuristics &  \tikzmark{c}& 23.1 & \textbf{66.5} & 30.3 & 14.8 & 34.1 & \textbf{451 ms}\\
+ Inter-branch information exchange & \tikzmark{d} & \textbf{23.9} & 66.0 & \textbf{31.2} & \textbf{15.5} & \textbf{35.0} & 484 ms\\
\end{tabular}
\begin{tikzpicture}[overlay, remember picture, shorten >=.5pt, shorten <=.5pt]
\draw [->]({pic cs:b}) [bend right] to ({pic cs:c}) node[above=5pt, left=1pt]{+2.1};
\draw [->]({pic cs:c}) [bend right] to ({pic cs:d}) node[above=5pt, left=1pt]{+0.8};
\end{tikzpicture}

\end{table*}

\begin{table*}[t]
\caption{The overall results of our method on the Cityscapes validation set.}
\label{tab:results_cityscapes_overall}
\centering
\begin{tabular}{ l r@{\hspace{2px}}|| c | c | c || c | c ||| c }
Method &  & PQ & SQ & RQ & PQ\textsubscript{Th} & PQ\textsubscript{St} & Prediction time\\
\hline
Multiple networks & \hspace{20px} & 43.7 & 74.5 & 55.6 & 36.5 & 48.9 & 1361 ms\\
Single network & \tikzmark{e}& 42.9 & 74.3 & 54.5 & 37.8 & 46.7 & 810 ms \\
+ Advanced heuristics &  \tikzmark{f}& 44.4 & 74.4 & 56.5 & 38.4 & 48.7 & \textbf{506 ms}\\
+ Inter-branch information exchange & \tikzmark{g} & \textbf{45.9} & \textbf{74.8} & \textbf{58.4} & \textbf{39.2} & \textbf{50.8} & 590 ms\\
\end{tabular}
\begin{tikzpicture}[overlay, remember picture, shorten >=.5pt, shorten <=.5pt]
\draw [->]({pic cs:e}) [bend right] to ({pic cs:f}) node[above=5pt, left=1pt]{+1.5};
\draw [->]({pic cs:f}) [bend right] to ({pic cs:g}) node[above=5pt, left=1pt]{+1.5};
\end{tikzpicture}

\end{table*}

\subsubsection{Overlap removal for things classes}
Because the instance segmentation prediction is essentially based on an object detector and many overlapping region proposals, there can be overlap between different predicted instance masks. In the baseline method proposed by \cite{Kirillov2018}, overlap is removed by prioritizing instance masks with higher corresponding classification scores. In our method, we choose to leverage the per-instance and per-pixel score maps to resolve conflicting sections. First, we transform all predicted instance masks to the full image size. Then, in the case that two or more instance masks predict that a certain pixel belongs to their object, the pixel is assigned to the instance mask with the highest score at that specific pixel. We choose to use per-pixel scores because it is more intuitive to solve per-pixel conflicts using per-pixel scores. As a result of this heuristic, all output pixels are assigned to only one object.

\subsubsection{Merging outputs from both branches}
\label{sec:method_heuristics_merging}
Unlike the \textit{stuff} classes, which are only considered in the semantic segmentation branch, the \textit{things} classes are part of the prediction of both the semantic segmentation and the instance segmentation branch. As a result, there are inevitably \textit{things} prediction conflicts between the two outputs. Because the semantic segmentation output does not distinguish between different instances of objects, the two outputs cannot be compared directly. Similarly to the baseline method in \cite{Kirillov2018}, we prioritize the instance segmentation output over the semantic segmentation output. In the baseline method, this is done by replacing all pixels with \textit{things} class predictions by the semantic segmentation branch with \textit{void} labels. To avoid the loss of potentially useful information, we improve the baseline heuristic by replacing \textit{void} labels by high scoring \textit{stuff} predictions, given that the score for that pixel is above a threshold $\alpha$. We use $\alpha = 0.25$. Finally, as in \cite{Kirillov2018}, the instance segmentation output is used to replace the \textit{stuff} and \textit{void} labels at pixels where it predicts \textit{things}. Because all these instance masks have a unique \textit{id}, the result of this heuristic is an output in the panoptic segmentation format.

\subsubsection{Removing unlikely \textit{stuff}}
As a third heuristic, any predicted \textit{stuff} class with a total pixel count below a given threshold is removed from the output as well. These predictions are then replaced by either \textit{void} labels or high scoring \textit{stuff} classes above this threshold, following the procedure described in Section \ref{sec:method_heuristics_merging}. This is done because it is very unlikely that a \textit{stuff} class consists of a small number of pixels, if it is present in an image. This heuristic is proposed by the organizers of the COCO Panoptic Segmentation Challenge during ECCV 2018, in their auxiliary code \cite{COCO2018}. In this code, they use a fixed pixel threshold of 4096. However, it is likely that this number depends on the size of the image. Therefore, we use a threshold that is a constant fraction, $f$, of the total amount of pixels of an image. The ideal value for this fraction depends on the dataset, as is described in Section \ref{sec:implementation_details}.

\section{IMPLEMENTATION DETAILS}
\label{sec:implementation_details}
We implement our methodology using TensorFlow. For training, we optimize the loss function in Eq. \ref{eq:loss} using a stochastic gradient descent optimizer with a momentum of 0.9. The loss and regularization weights are provided in Table \ref{tab:loss_weights}. These weights are found empirically and iteratively. Batch normalization is applied to all but the output layers, with a weight decay of 0.9. The network initialized using weights pre-trained on the ImageNet dataset \cite{Deng2009}, except for the models using inter-branch information exchange. When training these models, we initialize on a model that is pre-trained for semantic segmentation on the specific dataset, so that less unreliable semantic segmentation information is shared with the instance segmentation branch. We always use a single Nivia Titan Xp GPU for training.

\begin{table}[H]
\caption{The loss and regularization weights.}
\centering
\begin{tabular}{ c | c | c | c | c | c | c}
$\lambda_1$ & $\lambda_2$ & $\lambda_3$ & $\lambda_4$ & $\lambda_5$ & $\lambda_6$ & $\lambda_7$ \\ \hline
1.0 & 1.0 & 1.0 & 0.15 & 0.3 & 1.0 & 5.5e-5 \\
\end{tabular}
\label{tab:loss_weights}
\end{table}

We evaluate the network on two different street scene datasets: Cityscapes \cite{Cordts2016Cityscapes} and Mapillary Vistas \cite{Neuhold2017}. Because the two datasets have different properties, we use slightly different learning rate schedules and hyperparameters for training on each dataset.

\subsection{Cityscapes}
Cityscapes is a dataset that consists of 5k street scene images, which have all been taken in German cities. There are panoptic annotations for 8 \textit{things} classes and 11 \textit{stuff} classes. All images have a size of 1024 x 2048 pixels. For training, to allow for a batch size of 2, we resize the dimensions of the input images to 512 by 1024 pixels. For this dataset, we use a polynomial decay schedule for the learning rate, as in \cite{Chen2018}. We train for 30 epochs, and use an initial learning rate of 0.075 and a power of 0.9. Finally, it is found that stuff removal fraction $f = \frac{1}{512}$ leads to the best results for this dataset.

\subsection{Mapillary Vistas}
Mapillary Vistas is a more challenging dataset, consisting of 25k street scene images. The images have all been captured at different locations all around the world, and have panoptic annotations for 37 \textit{things} classes and 28 \textit{stuff} classes. The images have a very high resolution, the average being 2481 by 3419 pixels. To achieve state-of-the-art results on this method, high-resolution networks are required. The best-scoring instance segmentation method on Mapillary Vistas resizes input images so that the larger side is equal to 2400 pixels \cite{Liu2018}. However, this is not feasible in our implementation, because of memory requirements for joint learning and limited memory capacity. Therefore, the feature extractor has input dimensions of 640 x 900 pixels. This allows for the use of a batch size of 2. For Mapillary Vistas, we use stepwise learning rate schedule. We train for 21 epochs, use an initial learning rate 0.075, and multiply the learning rate by 0.5 after 8 and 14 epochs. Finally, for the stuff removal heuristic, we use a fraction of $f = \frac{1}{256}$.

\section{RESULTS}
\label{sec:results}
In this section, we present the results of our implemented network on Cityscapes and Mapillary Vistas. First, we describe the metrics in Section \ref{sec:results_metrics}. In addition to the overall performance of the network, discussed in Section \ref{sec:results_overall}, we also present ablation results on the the different inter-branch information exchange methods in Section \ref{sec:results_info_exchange}.

\subsection{Metrics}
\label{sec:results_metrics}
For panoptic segmentation evaluation, we use the Panoptic Quality (PQ) metric, as defined in \cite{Kirillov2018}. This PQ metric can be split into Segmentation Quality (SQ) and Recognition Quality (RQ), and is a product of these two terms. Here, the RQ indicates the ability of the network to recognize objects, and the SQ describes the ability to find accurate pixel masks for the objects that are actually detected. To investigate the performance of the two different branches of the network, we evaluate for \textit{things} (PQ\textsubscript{Th}) and \textit{stuff} (PQ\textsubscript{St}) classes separately as well. It should be noted that the range of scores achieved for the PQ metric varies heavily per dataset. It is not necessary for a network to achieve a score of 100 to be useful for self-driving vehicles applications. For instance, the top two pictures in Fig. \ref{fig:example_images} achieve a PQ score of 58.9 and 27.5, respectively.

To evaluate the real-time applicability of our system, we also evaluate the single image inference time when using a single Nvidia Titan Xp GPU. 

\begin{table}[t]
\caption{Results for different inter-branch information exchange methods, on the Mapillary Vistas validation set.}
\label{tab:results_info_exchange}
\centering
\begin{tabular}{ l ||@{\hspace{4px}} c @{\hspace{4px}}|@{\hspace{4px}} c @{\hspace{4px}}| @{\hspace{4px}}c @{\hspace{4px}}}
 Method & PQ & PQ\textsubscript{Th} & PQ\textsubscript{St} \\ \hline
Single network & 22.9 & 14.4 & 34.2 \\ 
+ Adding region proposals & 23.0 & 15.1 & 33.5 \\
+ Expanding bounding boxes & 23.1 & 15.3 & 33.5  \\
+ Implicit information exchange & \textbf{23.9} & \textbf{15.5} & \textbf{35.0}\\
\end{tabular}
\vspace{-15pt}
\end{table}

\vspace{-1pt}

\subsection{Overall results}
\label{sec:results_overall}
The overall results for the Mapillary Vistas and Cityscapes datasets are presented in Table \ref{tab:results_mapillary_overall} and \ref{tab:results_cityscapes_overall}, respectively. Firstly, we compare the single network with the approach using separate networks, using the baseline heuristics from \cite{Kirillov2018}. It can be seen that jointly learning the tasks in a single network greatly reduces the required prediction time. However, there is a drop in performance on the PQ metric. Also, it should be noted that the prediction time can be reduced even further by optimizing the implementation of the model for speed. This has not been done for our implementation.

With respect to the baseline single network, we first improve the performance by using advanced heuristics. It can be seen that this especially improves the performance of the \textit{stuff} classes. This is as expected, since most of the improvements to the heuristics aimed at making more accurate \textit{stuff} predictions. Moreover, it is found that the prediction time is decreased as well. This is the result of implementing the new overlap removal heuristic, that directly compares the per-pixel scores. Secondly, implementing the inter-branch information exchange gives a final performance boost. Ablation results on the inter-branch information exchange are presented in Section \ref{sec:results_info_exchange}. 

\begin{figure}[t]
\centering
\includegraphics[width=0.48\linewidth]{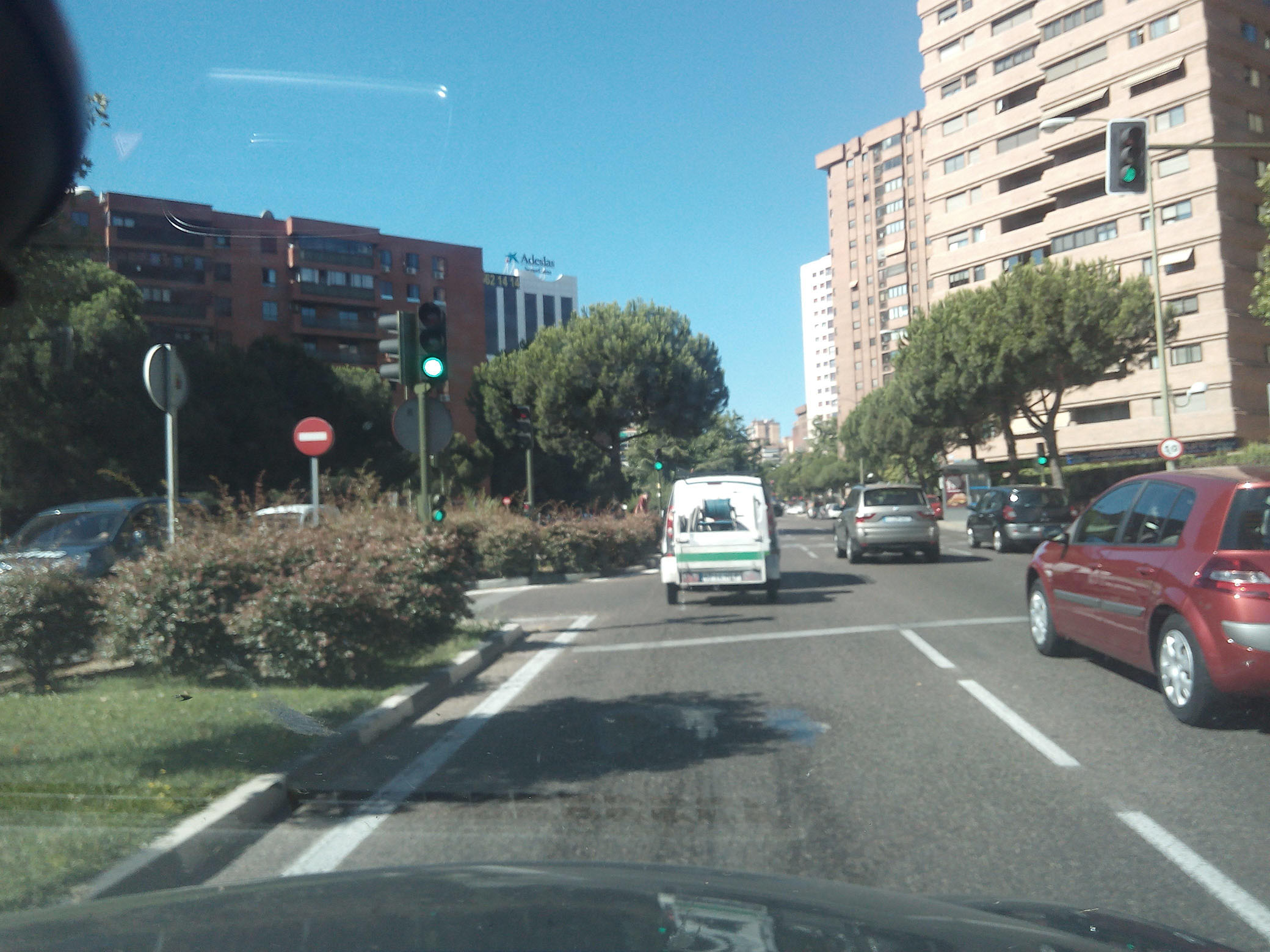}
\includegraphics[width=0.48\linewidth]{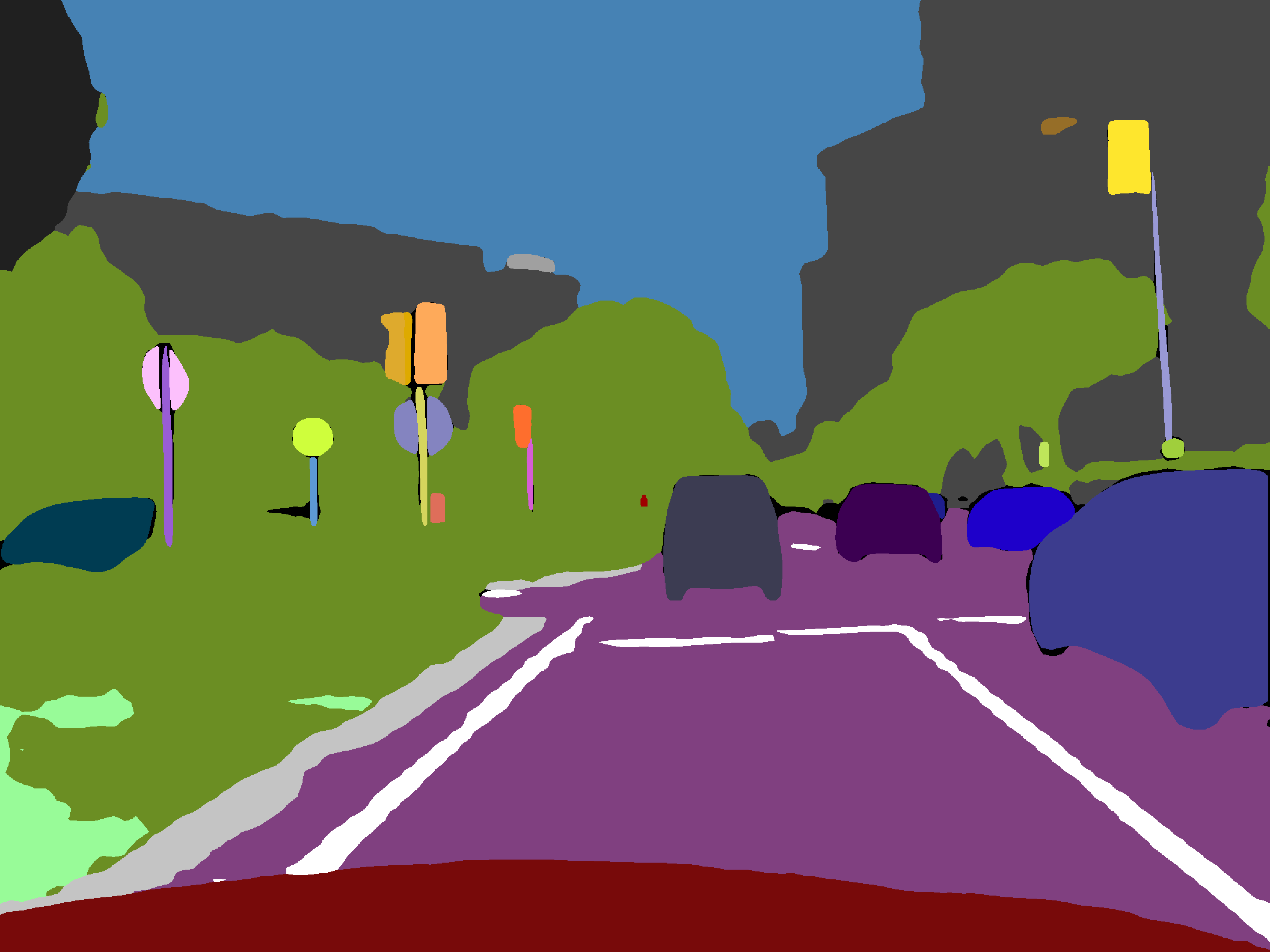}\\
\includegraphics[width=0.48\linewidth]{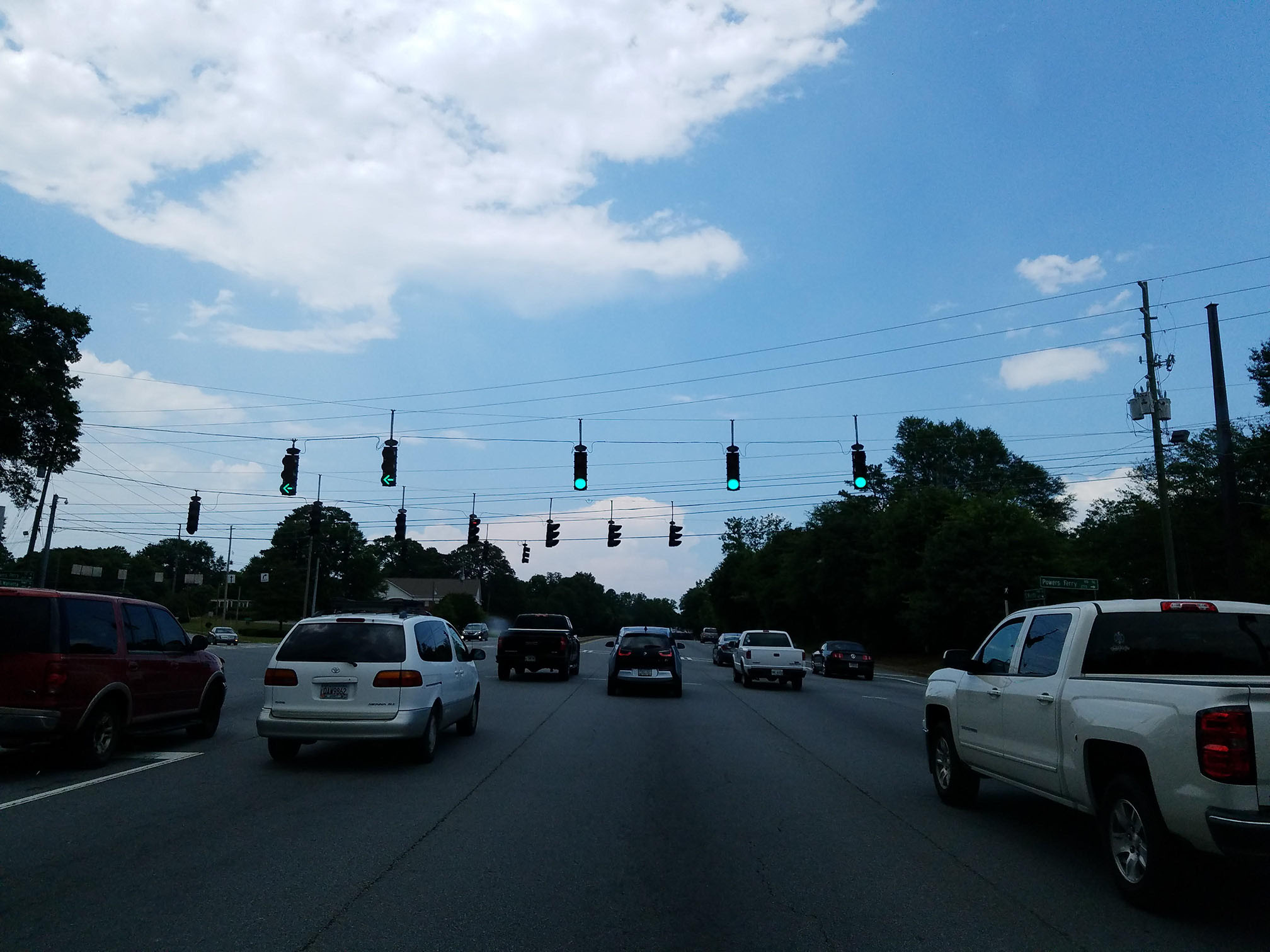}
\includegraphics[width=0.48\linewidth]{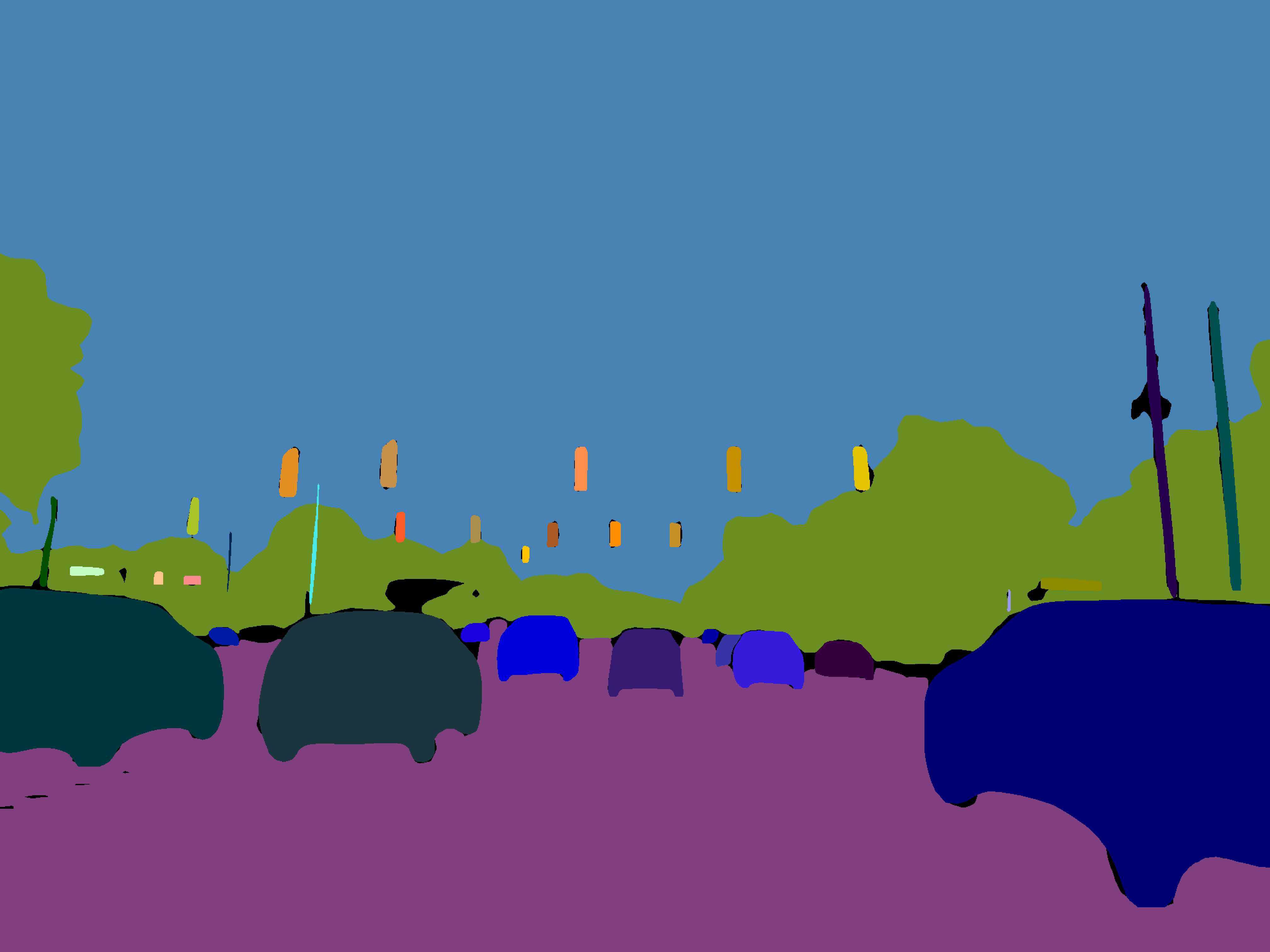}\\
\includegraphics[width=0.48\linewidth]{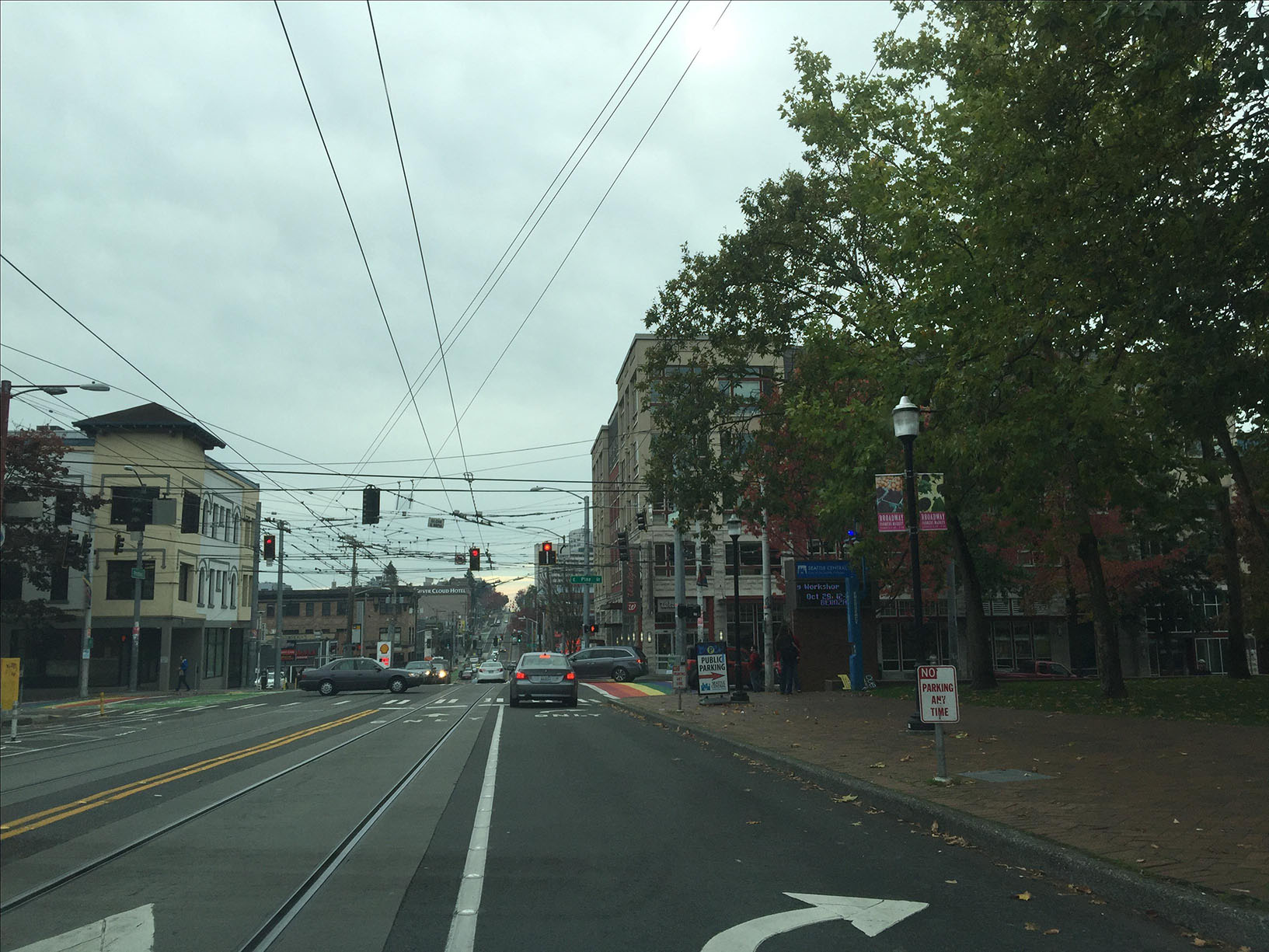}
\includegraphics[width=0.48\linewidth]{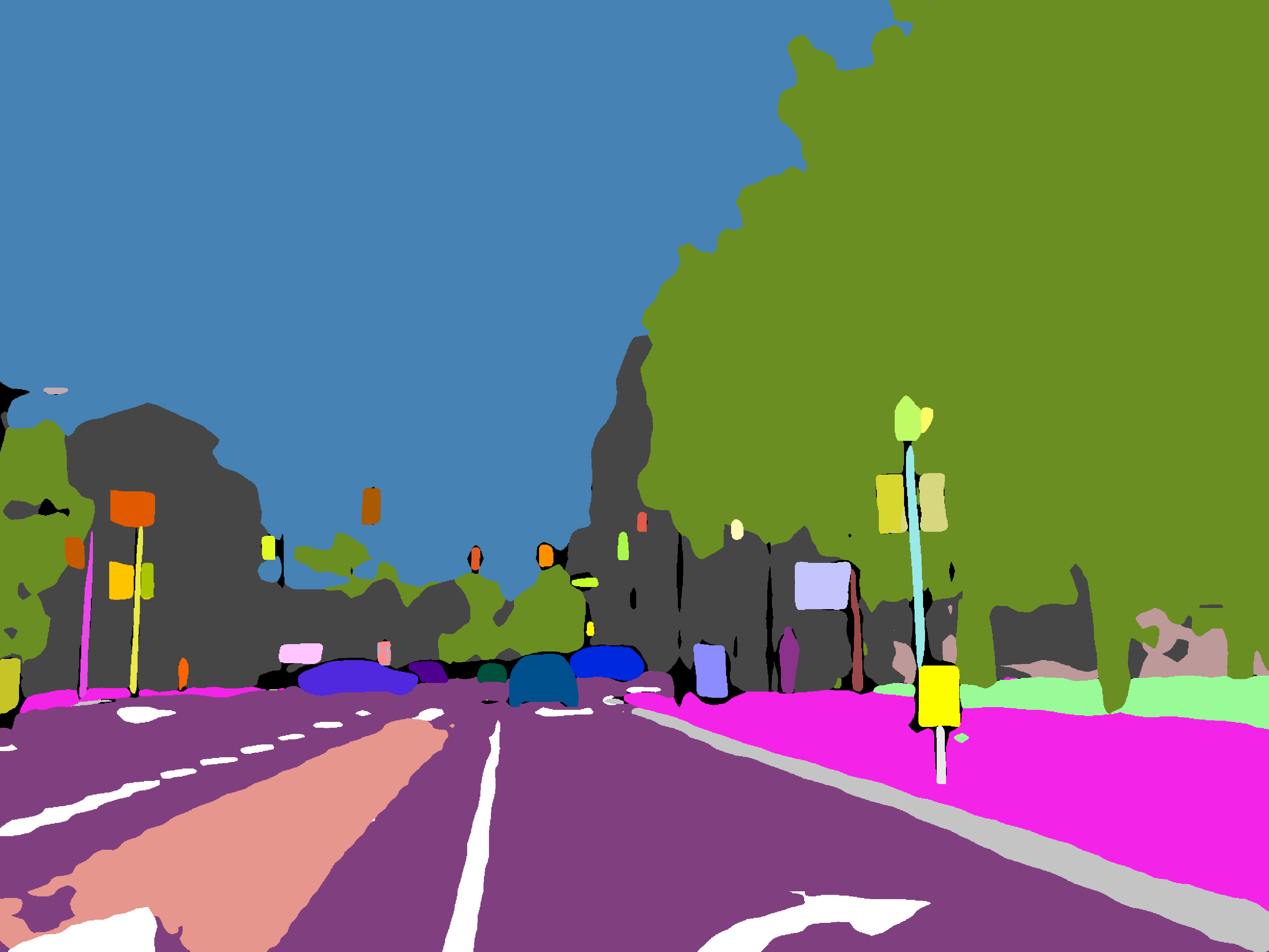}\\\includegraphics[width=0.48\linewidth]{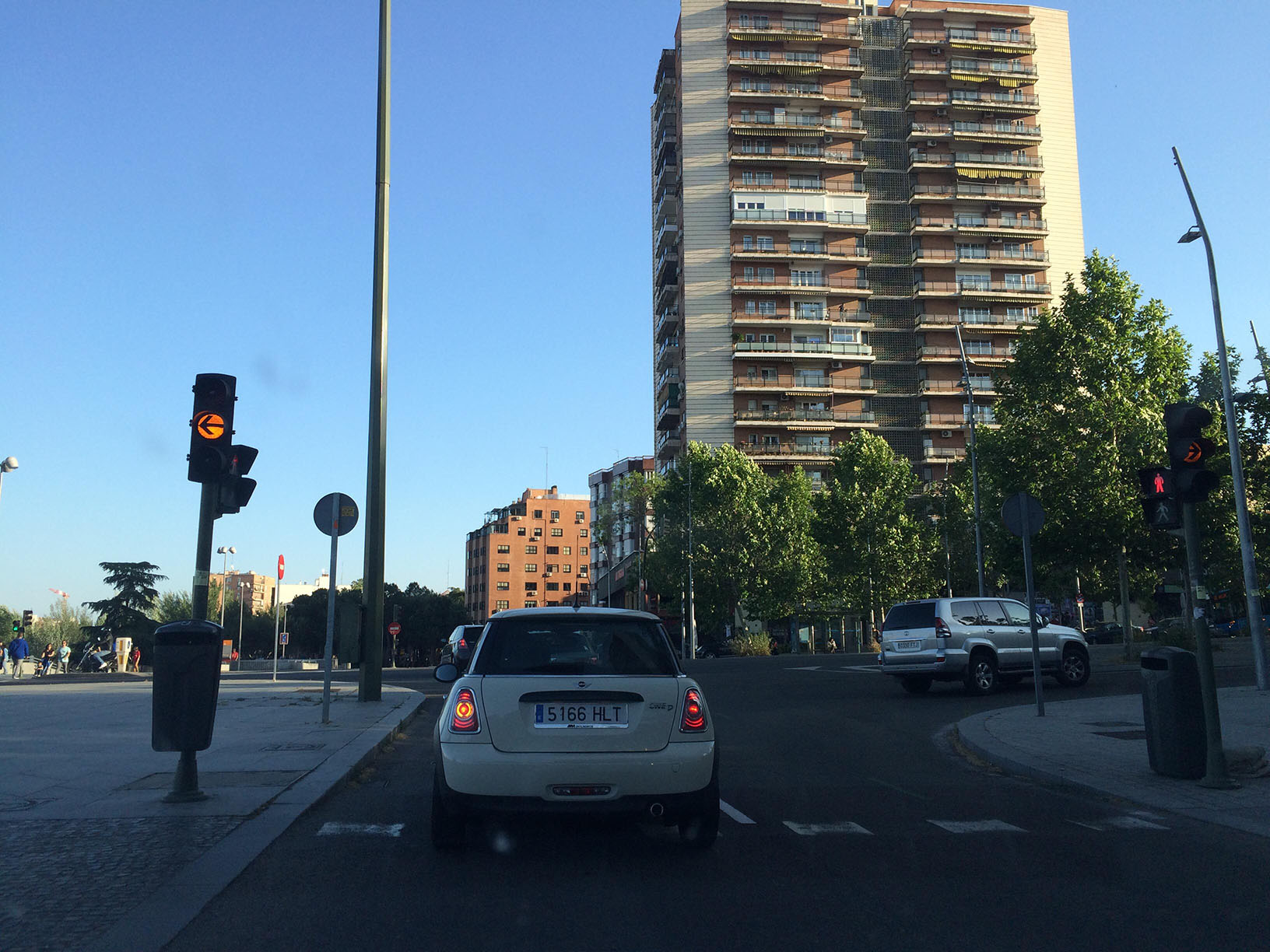}
\includegraphics[width=0.48\linewidth]{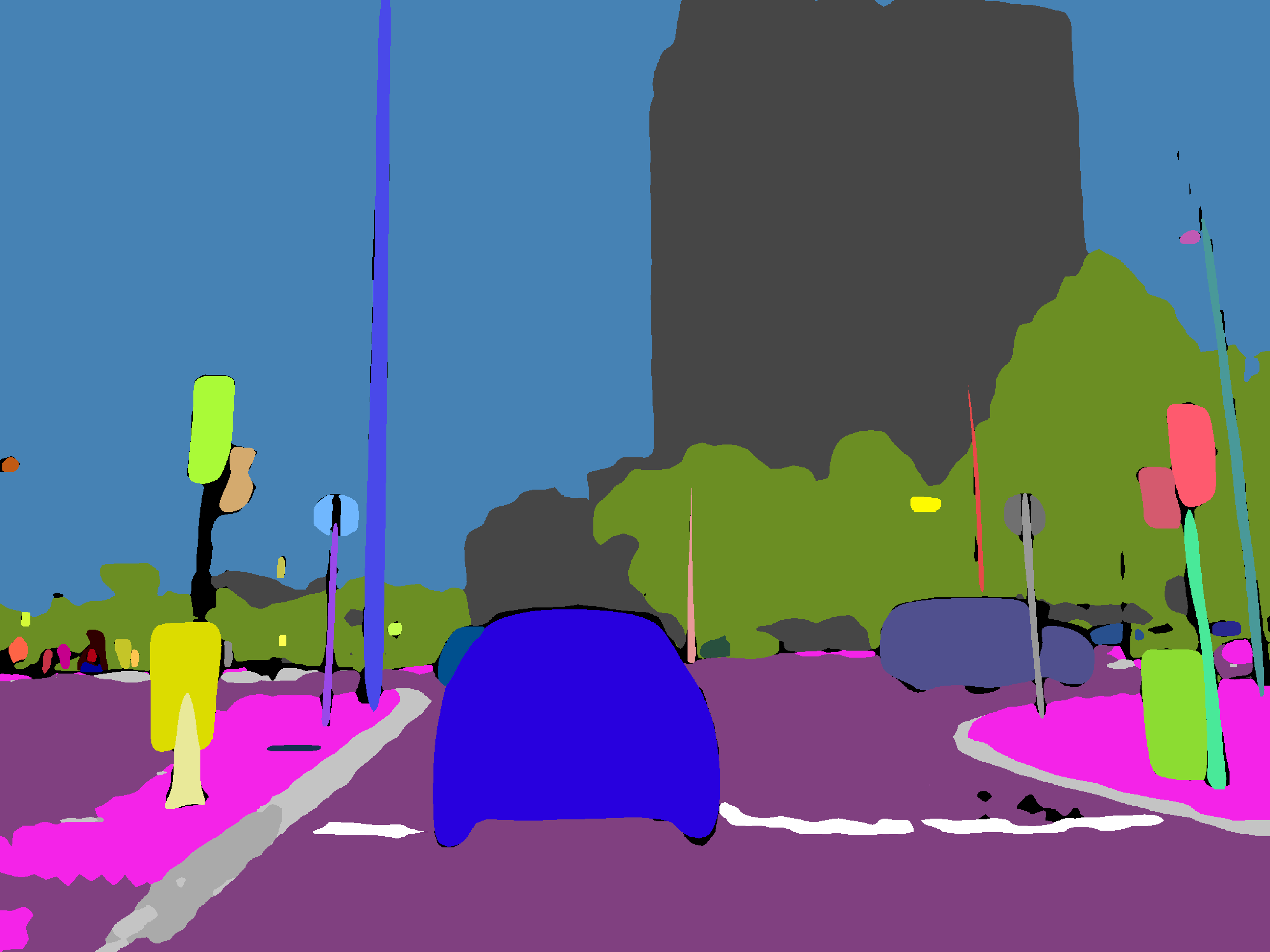}\\
\caption{Panoptic segmentation predictions by the network. Images from the Mapillary Vistas validation set. Each output pixel receives a color-coded class label and specific objects also receive a color-coded identifier label.} 
\label{fig:example_images}
\vspace{-10pt}
\end{figure}

\begin{figure*}[t]
\centering
    \centering
\subfloat{\includegraphics[width=.245\linewidth]{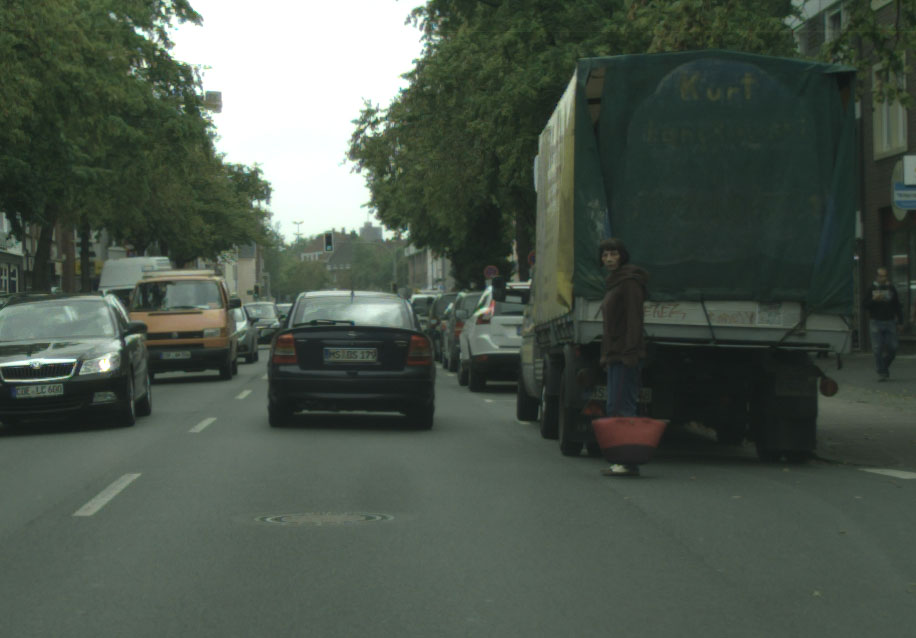}}
    \hfill
\subfloat{\includegraphics[width=.245\linewidth]{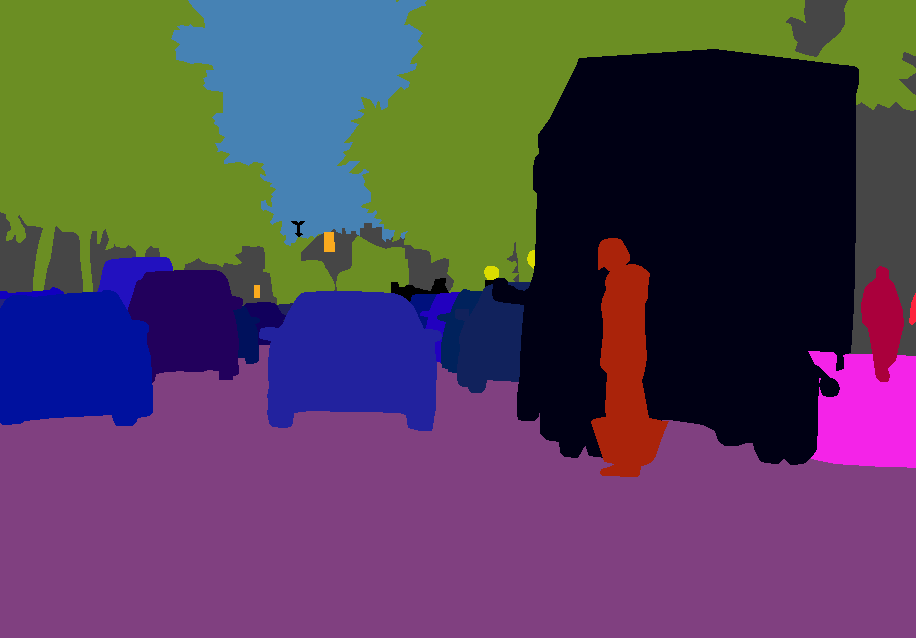}}
	\hfill
\subfloat{\includegraphics[width=.245\linewidth]{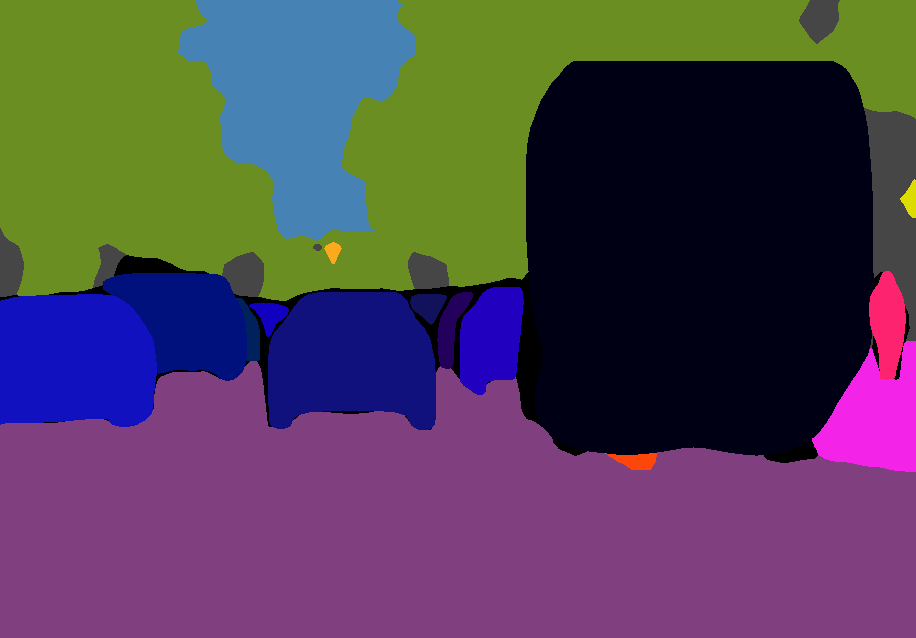}}
    \hfill
\subfloat{\includegraphics[width=.245\linewidth]{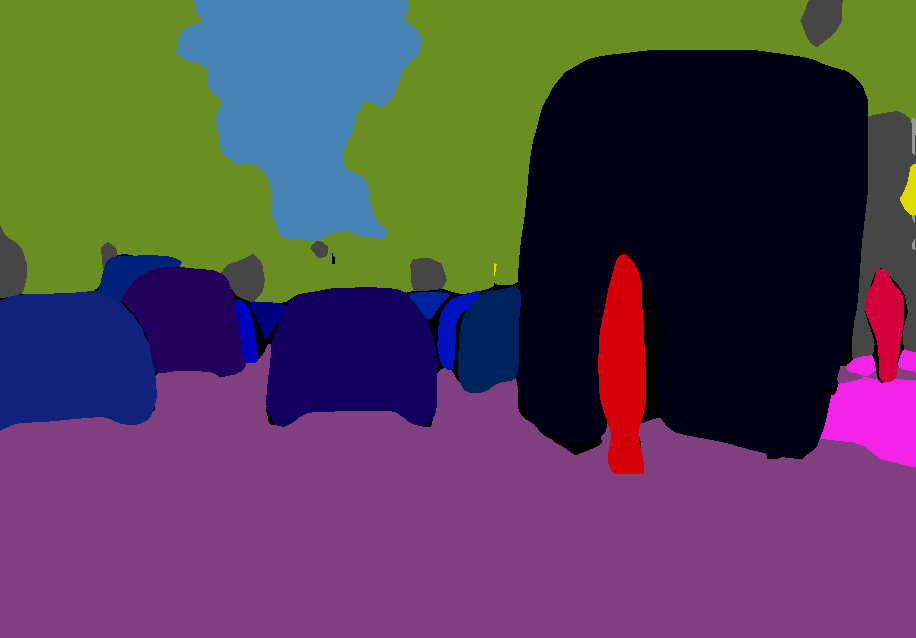}}\\
\subfloat[Input image]{\includegraphics[width=.245\linewidth]{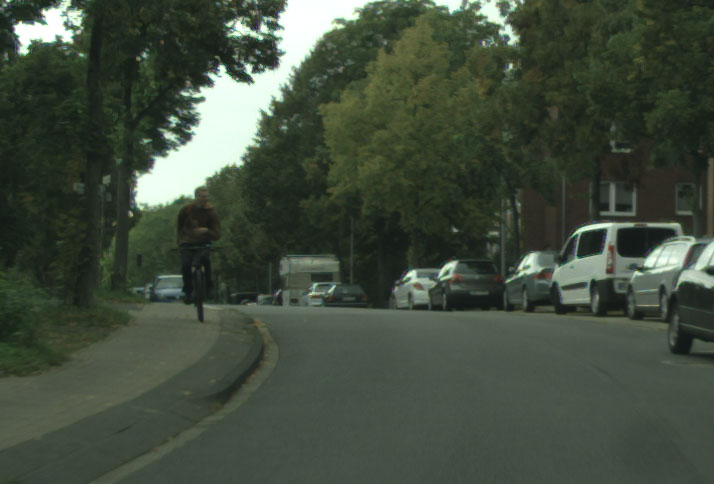}}
    \hfill
\subfloat[Ground truth]{\includegraphics[width=.245\linewidth]{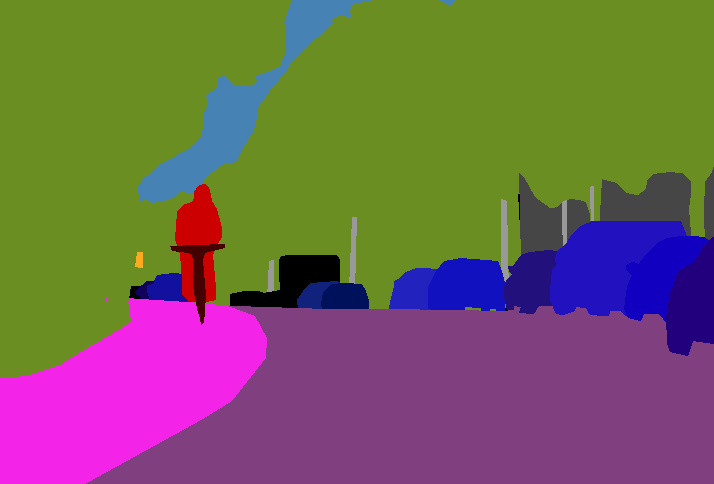}}
	\hfill
\subfloat[Baseline separate networks]{\includegraphics[width=.245\linewidth]{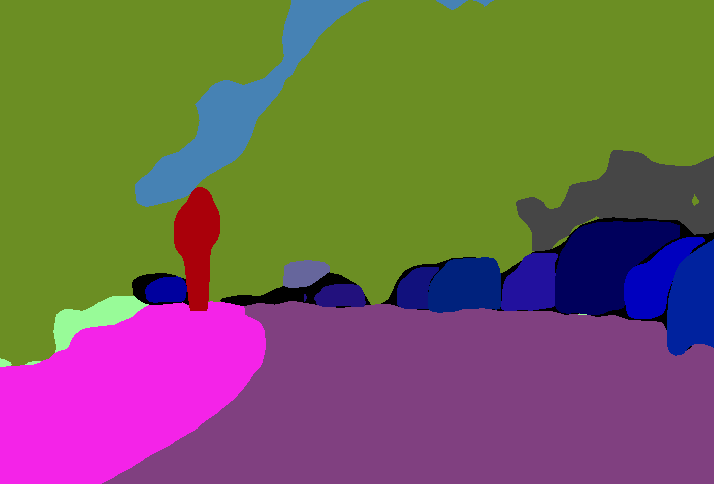}}
    \hfill
\subfloat[Our single network]{\includegraphics[width=.245\linewidth]{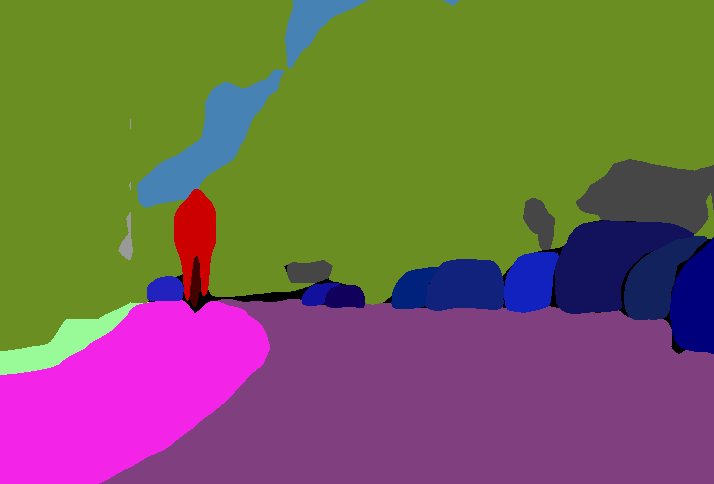}}\\

\caption{Panoptic segmentation examples on crops from the Cityscapes validation set. Different color shades indicate different instances of a certain class, and are randomly generated. Our single network is able to detect the pedestrian and bicycle due to the information flow between the branches and the advanced heuristics.}
\label{fig:examples_cityscapes}
\end{figure*}

Qualitative results of our method are shown in Fig. \ref{fig:example_images}. In Fig. \ref{fig:examples_cityscapes}, we compare predictions by our network with predictions by the separate networks and the ground truth.

\subsection{Inter-branch information exchange ablation results}
\label{sec:results_info_exchange}
Inter-branch information exchange is the final contribution of our method. In Table \ref{tab:results_mapillary_overall}, it has already been shown that the information exchange improves the overall performance on the Mapillary Vistas dataset. In Table \ref{sec:results_info_exchange}, we evaluate the performance of the different individual information exchange methods. Note that all methods in this table are initialized on a pre-trained semantic segmentation model, for a fair comparison.  The results show that the addition of region proposals and expansion of bounding boxes improves the PQ score on the \textit{things} classes, as intended. The additional implicit information exchange between the branches impacts the performance of both branches, because of the additional information flow passing through both branches.

In Table \ref{tab:results_cityscapes_overall}, it can be seen that including inter-branch information exchange also improves the performance of the network on the Cityscapes dataset. Again, the PQ is improved on both the \textit{things} and the \textit{stuff} classes, while only slightly increasing the required prediction time.

%

\section{CONCLUSIONS}
\label{sec:conclusions}
With this work, we have taken a step towards holistic street scene understanding from image data, by presenting a single deep neural network for panoptic segmentation. This single network approach allows for easier implementation on devices, such as intelligent vehicles. Moreover, it reduces the needed computation time by a factor of 2 with respect to the use of separate networks. It is shown that, for a single network approach to achieve better Panoptic Quality than separately learned networks, it is crucial to exchange additional information between different parts of the single network. Moreover, we improve the merging heuristics by using the most likely and most reliable information from both the instance segmentation and semantic segmentation outputs. These improvements result in a performance increase of +2.9 and +3.0 on the PQ metric, on the Mapillary Vistas and Cityscapes validation sets, respectively. In future work, to realize full end-to-end panoptic segmentation, our aim is to research a differentiable merging method to replace the current non-differentiable merging heuristics.

\addtolength{\textheight}{-12cm}   



%
%
%
%


{\small
\bibliographystyle{ieeetran}
\bibliography{bibliography}

\begin{thebibliography}{10}
\bibitem{CodeGitlab2019}
The code will be made publicly available and the URL will be inserted after
  acceptance.
\end{thebibliography}
}

\end{document}